\documentclass[]{fairmeta}

\usepackage[utf8]{inputenc}
\usepackage[T1]{fontenc}
\usepackage{geometry}
\usepackage{hyperref}
\usepackage{graphicx}
\usepackage{amsmath}
\usepackage{amssymb}
\usepackage{enumitem}
\usepackage{booktabs}
\usepackage{xcolor}
\usepackage{tcolorbox}
\usepackage{tikz}
\usepackage{fontawesome5}
\usepackage{evaluationcard3}

\definecolor{metablue}{RGB}{24,119,242}
\definecolor{metadark}{RGB}{101,103,107}

\author[1]{Florian Bordes}
\author[1]{Candace Ross}
\author[1]{Justine T Kao}
\author[1]{Evangelia Spiliopoulou}
\author[1]{Adina Williams}
\affiliation[1]{FAIR at Meta}

\newcommand{\cardName}{\texttt{Eval Factsheet}}
\newcommand{\cardNamePlural}{\texttt{Eval Factsheets}}
\newcommand{\cardNameFull}{\texttt{Eval Factsheets (EFS)}}

\title{\textbf{\cardNamePlural: A Structured Framework for\\Documenting AI Evaluations}}

\abstract{The rapid proliferation of benchmarks has created significant challenges in reproducibility, transparency, and informed decision-making. However, unlike datasets and models---which benefit from structured documentation frameworks like Datasheets and Model Cards---evaluation methodologies lack systematic documentation standards. We introduce \cardNameFull, a structured, descriptive framework for documenting AI system evaluations through a comprehensive taxonomy and questionnaire-based approach. Our framework organizes evaluation characteristics across five fundamental dimensions: \textbf{Context} (Who made the evaluation and when?), \textbf{Scope} (What does it evaluate?), \textbf{Structure} (With what the evaluation is built?), \textbf{Method} (How does it work?) and \textbf{Alignment} (In what ways is it reliable/valid/robust?). We implement this taxonomy as a practical questionnaire spanning five sections with mandatory and recommended documentation elements. Through case studies on multiple benchmarks, we demonstrate that \cardNameFull\ effectively captures diverse evaluation paradigms---from traditional benchmarks to LLM-as-judge methodologies---while maintaining consistency and comparability.
We hope \cardNamePlural\ are incorporated into both existing and newly released evaluation frameworks and lead to more transparency and reproducibility.
}

\date{\today}
\metadata[Code]{\url{https://github.com/facebookresearch/EvalFactsheets}}

\begin{document}

\maketitle

\section{Introduction}
\label{sec:intro}

Self-supervised learning, large language models, and multimodal systems have achieved remarkable performance across diverse tasks, yet a critical challenge persists: evaluating these systems requires methodologies as advanced as the models themselves. A single state-of-the-art model might be assessed through hundreds of benchmarks spanning multiple modalities, evaluation paradigms, and fairness or safety criteria. Despite this proliferation of benchmarks, evaluation methodologies lack the systematic documentation standards that have become common for datasets~\citep{gebru2021datasheets} and models~\citep{mitchell2019model}, creating opacity in how AI systems are actually assessed and compared.

This documentation gap for evaluation manifests in three concrete challenges. First, evaluation methodologies often embed hidden assumptions about data distributions, access paradigms, and validity conditions. Because these assumptions are rarely explicitly stated, it can lead to misapplication and misinterpretation of results. Second, the absence of standardized reporting prevents meaningful comparison across evaluations: two benchmarks claiming to measure ``reasoning capability'' may implement this construct in fundamentally incompatible ways. Third, reproducibility suffers when critical details---e.g. judge selection, contamination checks, statistical validation procedures---remain undocumented or described inconsistently across papers.

Existing documentation frameworks address adjacent but distinct concerns. Datasheets for Datasets~\citep{gebru2021datasheets} document data provenance and composition, but not the evaluation methodologies applied to that data. Model Cards~\citep{mitchell2019model} document system capabilities and limitations, but treats evaluation as a secondary component. Recent comprehensive evaluation frameworks like HELM~\citep{liang2022helm} and BIG-bench~\citep{srivastava2022beyond} provide extensive benchmark suites, but focus on breadth of coverage rather than documentation standards. While these efforts have improved AI transparency, none provides a general-purpose framework specifically designed for documenting evaluation methodologies themselves.

To address this gap, we introduce \cardNameFull, a structured framework for documenting AI evaluation methodologies through a systematic taxonomy and questionnaire-based approach. The key insight behind \cardNamePlural\ is that evaluation methodologies, despite their diversity, share fundamental characteristics that can be organized into five orthogonal dimensions: \textbf{Context} (Who made the evaluation and when?), \textbf{Scope} (What does it evaluate?), \textbf{Structure} (With what the evaluation is built?), \textbf{Method} (How does it work?) and \textbf{Alignment} (In what ways is it reliable/valid/robust?). We provide the detailed taxonomy in \Cref{sec:taxonomy} and how we operationalize the taxonomy into the \cardNamePlural\ framework in \Cref{sec:framework}. Lastly, in \Cref{sec:examples}, we demonstrate the framework's applicability through diverse case studies. Across these cases, the framework maintains consistent structure while accommodating paradigm-specific requirements.

Our main contributions are:

\begin{enumerate}
\item \textbf{A five-dimensional taxonomy} organizing evaluation characteristics across Context, Scope, Structure, Method and Alignment, providing a systematic framework for categorizing evaluation methodologies.

\item \textbf{A structured questionnaire} to easily generate \cardNamePlural, implementing 27 questions that balance mandatory requirements with flexible optional elements to accommodate diverse evaluation types.

\item \textbf{Comprehensive case studies} on widely used benchmarks (ImageNet, HumanEval, MT-Bench) demonstrating that the framework effectively captures evaluations spanning traditional benchmarks, execution-based testing, and LLM-as-judge paradigms.

\item \textbf{Practical resources} including templates, HTML forms, completion guides, and integration recommendations for adoption in research and deployment contexts.
\end{enumerate}

\section{Related Work}
\label{sec:related}

Our work builds on a growing body of research advocating for systematic documentation of machine learning artifacts. We position \cardNamePlural\ within this broader ecosystem and discuss their relationship to evaluation methodology research.

\subsection{Documentation Frameworks for AI Systems}

Several frameworks address AI system documentation. Datasheets for Datasets~\citep{gebru2021datasheets} introduced structured questionnaires covering data provenance, composition, and uses, establishing the template our work adapts for evaluations. Building on this foundation,~\citet{bender2018data} developed Data Statements for NLP datasets emphasizing demographic representation, while ~\cite{pushkarna2022data} created Data Cards for TensorFlow datasets with interactive visualization. ~\cite{holland2020dataset} proposed Dataset Nutrition Labels providing at-a-glance summaries of dataset characteristics.

Model Cards~\citep{mitchell2019model} document trained model characteristics, intended uses, and performance across demographic groups, emphasizing fairness considerations.~\cite{arnold2019factsheets} developed FactSheets for enterprise AI systems, providing comprehensive technical specifications alongside accountability considerations. More recently, System Cards \citep{gursoy2022system} have emerged to document complete AI systems holistically, including training data, model architecture, and deployment context.

While these frameworks have significantly improved AI transparency, they treat evaluation as a component to be reported rather than a methodology requiring independent documentation. Model Cards, for instance, include performance metrics but provide no structured approach for documenting how those metrics were obtained, what assumptions underlie their interpretation, or how to compare across different evaluation methodologies. \cardNamePlural\ fills this gap by providing evaluation-specific documentation with the same rigor previously applied to datasets and models.

\subsection{Comprehensive Evaluation Frameworks}

Recent efforts have developed comprehensive benchmarks evaluating models across multiple dimensions. HELM~\citep{liang2022helm} assesses language models with emphasis on standardized evaluation protocols. VHELM \citep{lee2024vhelm} similarly focuses on a standardized taxonomy, expanding to vision-language models. These taxonomies of evaluation scenarios share conceptual similarities with our evaluation taxonomy, though they focus on (vision-)language model capabilities rather than evaluation documentation. BIG-bench~\citep{srivastava2022beyond} crowdsources tasks targeting behaviors believed beyond current model capabilities, emphasizing diversity in evaluation. GEM~\citep{gehrmann2022gem} provides evaluation frameworks for natural language generation with detailed annotation protocols and multiple evaluation paradigms.

These frameworks make important contributions to evaluation breadth and standardization but focus on benchmark creation rather than evaluation documentation. HELM provides detailed methodology for its own evaluations but not a generalizable framework for documenting arbitrary evaluation methodologies. Our work is complementary: while HELM standardizes how to conduct certain evaluations, \cardNamePlural\ standardizes how to document any evaluation, including HELM itself. This distinction allows our framework to span traditional benchmarks, execution-based testing, human evaluation, and emerging paradigms like LLM-as-judge.

\subsection{Evaluation Methodology and Meta-Evaluation}

Growing attention to evaluation methodology has identified critical challenges. Benchmark saturation and contamination threaten validity as models are increasingly trained on test data~\citep{sainz2023nlp,brown2020language}. Construct validity---whether evaluations measure what they claim---remains underexamined~\citep{raji2021ai}. LLM-as-judge methodologies introduce new considerations around evaluator bias and agreement with human judgment~\citep{mtbench}.

Our taxonomy directly addresses these concerns through dedicated dimensions. The Alignment category systematically organizes contamination risks, validity threats, and sources of variance while the Method category (\S\ref{subsec:method}) provides structured documentation for emerging paradigms including model-based judges. By operationalizing these meta-evaluation concerns into concrete documentation requirements, we transform methodological critiques into actionable improvements.

\subsection{Domain-Specific and Reproducibility Efforts}

Reproducibility initiatives have developed checklists and reporting standards for machine learning research~\citep{pineau2021improving,dodge2019show}. Domain-specific guidelines address unique evaluation challenges in medical AI~\citep{wiens2019diagnostic}, autonomous systems, and fairness assessment~\citep{mitchell2021algorithmic}. While valuable for their domains, these efforts lack a unified framework spanning evaluation types.

\cardNamePlural\ provides this generalization while maintaining compatibility with domain-specific requirements. Our questionnaire includes optional sections that can be extended with domain-specific elements, and our taxonomy's flexibility accommodates specialized evaluation paradigms. Rather than replacing domain standards, \cardNamePlural\ offer a common foundation that domain guidelines can build upon, creating an interconnected ecosystem of ML documentation.

\section{Our Evaluation Taxonomy}
\label{sec:taxonomy}

We propose a taxonomy that decomposes any evaluation along five orthogonal dimensions: \textbf{Context} addresses who creates the evaluation and when; \textbf{Scope} defines what properties and capabilities are measured; \textbf{Structure} specifies the data sources and organization used to build the evaluation; \textbf{Method} describes the operational procedure for conducting evaluation; and \textbf{Alignment} encompasses reliability, validity, and robustness concerns. This decomposition enables consistent documentation across evaluation paradigms ranging from traditional supervised benchmarks to emerging LLM-as-judge arenas.

Our taxonomy emerged from analyzing hundreds of existing evaluations across computer vision, natural language processing, speech recognition, multimodal systems, and reinforcement learning domains. We identified recurring patterns in how evaluations are designed, validated, and deployed, which we distill into these five fundamental dimensions. The following subsections detail each dimension with formal definitions and concrete examples illustrating the distinctions.

\subsection{Context: Who and When}
\label{subsec:context}

The \textbf{Context} dimension captures the provenance and purpose of an evaluation---who created it, when it was released, and why it exists. Understanding evaluation context is crucial because the same measurement approach may be appropriate for one purpose but inadequate for another.

\paragraph{Provenance and Attribution.} Evaluation provenance establishes accountability and expertise claims. We document the \textit{organization, university, company, laboratory, or research group} responsible for the evaluation, along with specific \textit{authors or contributors}. This attribution enables assessment of domain expertise, potential conflicts of interest, and institutional backing. The \textit{release date} provides temporal context, as evaluation standards evolve over time---what constituted rigorous evaluation in 2018 may be inadequate by 2024 standards. Associated \textit{research papers} and \textit{code repositories} provide detailed methodology and enable replication.

\paragraph{Purpose and Stakeholders.} We identify four primary evaluation purposes, each serving distinct stakeholders with different requirements:

\begin{itemize}
    \item \textit{Development-focused} evaluations guide model training and improvement, emphasizing rapid iteration, diagnostic granularity, and actionable feedback. These serve researchers and engineers actively developing systems, requiring fast execution and detailed error analysis over absolute performance numbers.
    
    \item \textit{Selection-focused} evaluations help choose between competing models for specific applications. Decision-makers require robust comparison capabilities, clear performance differentials, and alignment between evaluation tasks and deployment scenarios.
    
    \item \textit{Deployment-focused} evaluations assess production readiness, often emphasizing safety, robustness, and edge-case behavior. Operators and regulators require comprehensive coverage of failure modes, with particular attention to rare but consequential errors.
    
    \item \textit{Research-focused} evaluations advance scientific understanding of model capabilities and limitations. The broader research community requires rigorous construct validity, reproducibility, and generalizability over immediate practical utility.
\end{itemize}

The same evaluation may serve multiple purposes, but design priorities often conflict. Development evaluations favor speed and diagnostic detail; selection evaluations favor standardization; deployment evaluations favor conservatism; research evaluations favor theoretical grounding. Explicitly documenting intended purpose helps users assess appropriateness for their needs.

\subsection{Scope: What Is Measured}
\label{subsec:scope}

The \textbf{Scope} dimension defines what specific properties, capabilities, and principles an evaluation assesses. Clear scope specification prevents misinterpretation of evaluation results and enables meaningful comparison across different measurement approaches.

\paragraph{Capabilities and Principles.} Evaluations target specific \textit{capabilities}---cognitive or functional competencies that models may possess. Examples include intuitive physics (understanding object permanence and motion), common sense reasoning (leveraging implicit world knowledge), mathematical problem-solving (manipulating formal systems), or creative generation (producing novel coherent outputs). These capabilities often map to cognitive science constructs or practical task requirements. We distinguish capabilities (what models can do) from the data modalities used to test them (how we probe those capabilities).

\paragraph{Model Properties.} Beyond task-specific capabilities, evaluations assess broader \textit{model properties} that cut across domains:

\begin{itemize}
    \item \textit{Performance}: Fundamental task accuracy, precision, recall, or task-specific metrics reflecting primary objectives.
    
    \item \textit{Quality}: Multi-faceted quality assessment including fluency, coherence, factuality, relevance, or domain-specific quality criteria.
    
    \item \textit{Robustness}: Performance stability under perturbations, distribution shifts, or adversarial manipulation.
    
    \item \textit{Calibration}: Alignment between model confidence and actual accuracy, crucial for decision-making applications.
    
    \item \textit{Adversarial Robustness}: Performance on inputs specifically designed to elicit failures, testing security and reliability.
    
    \item \textit{Memorization}: Distinguishing genuine learned patterns from memorized training examples, important for privacy and generalization.
    
    \item \textit{Fairness}: Equitable performance across demographic groups, protected attributes, or population segments.
    
    \item \textit{Safety}: Avoidance of harmful, toxic, or dangerous outputs across diverse prompt types.
    
    \item \textit{Leakage and Contamination}: Detection of inappropriate influence from evaluation data appearing in training.
    
    \item \textit{Privacy}: Protection of sensitive information, both in training data and user interactions.
    
    \item \textit{Interpretability}: Understandability and explainability of model decisions and internal representations.
    
    \item \textit{Efficiency}: Computational, memory, and energy requirements for training and inference.
    
    \item \textit{Retrainability}: Ease of updating models with new data or adapting to evolving requirements.
    
    \item \textit{Meta-Learning}: Adaptation speed to new tasks or domains with limited examples.
\end{itemize}

\paragraph{Modality Specification.} The \textit{input modality} defines what the model receives (text, vision, audio, video, code, structured data, or multimodal combinations), while the \textit{output modality} defines what the model produces (text, generated images, audio, code, actions, embeddings, or multimodal combinations). The same capability (e.g., reasoning) may be tested through different modality combinations (text-to-text logical puzzles versus vision-to-text visual reasoning). Modality choice affects accessibility (what systems can be evaluated), difficulty (how challenging measurement becomes), and ecological validity (how well evaluation reflects real usage).

Comprehensive scope documentation enables users to understand exactly what claims an evaluation supports and what falls outside its purview. An evaluation may rigorously measure mathematical problem-solving (narrow scope) without saying anything about creative writing, multimodal reasoning, or robustness (excluded from scope).

\subsection{Structure: Data Composition and Organization}
\label{subsec:structure}

The \textbf{Structure} dimension describes how evaluation data is sourced, organized, and maintained. Structural choices profoundly affect evaluation validity, difficulty, and susceptibility to contamination.

\paragraph{Input Data Sources.} Evaluation inputs may derive from multiple sources, each with distinct implications:

\begin{itemize}
    \item \textit{Existing datasets}: Leveraging established benchmarks (e.g. MS COCO, ImageNet, Wikipedia, Common Crawl, GitHub repositories) enables comparison with prior work but risks contamination as these datasets appear in training corpora.
    
    \item \textit{New datasets}: Purpose-built evaluation data released with the benchmark provides control over difficulty and content but requires significant annotation effort.
    
    \item \textit{Proprietary/closed datasets}: Private data prevents contamination but limits reproducibility and community participation.
    
    \item \textit{Synthetic/generated data}: Programmatically or procedurally generated inputs enable unlimited scale and precise control over properties.
    
    \item \textit{Crowdsourced data}: Contributed by non-expert users, providing diversity but requiring quality control.
    
    \item \textit{Expert-curated data}: Domain specialists create challenging, high-quality examples but at significant cost.
    
    \item \textit{Real-world deployment data}: Actual usage examples provide ecological validity but raise privacy concerns and may be unrepresentative.
\end{itemize}

\paragraph{Output Reference Sources.} Ground truth or reference outputs against which models are compared may originate from:

\begin{itemize}
    \item \textit{Human annotations}: Expert or crowdsourced labels provide authoritative references but introduce annotator variance and potential biases.
    
    \item \textit{Existing dataset labels}: Inherited from source datasets, providing consistency with prior work but potentially propagating annotation errors.
    
    \item \textit{Programmatic generation}: Rule-based or simulation-derived ground truth offers perfect consistency but may oversimplify complex phenomena.
    
    \item \textit{Execution-based verification}: For code or action outputs, functional correctness determined by interpreters or simulators provides objective grounding.
    
    \item \textit{Model-generated references}: Using strong models to generate references enables scale but introduces model biases and potential circularity.
    
    \item \textit{Reference-free evaluation}: Some properties (diversity, fluency) can be assessed without explicit ground truth, though validity concerns intensify.
\end{itemize}

\paragraph{Size and Scale.} Evaluation size profoundly affects statistical power and practical feasibility. We categorize:

\begin{itemize}
    \item \textit{Small} (< 1K samples): Enables deep analysis but provides limited statistical power; common for expensive expert evaluation.
    
    \item \textit{Medium} (1K--100K samples): Balances coverage and feasibility; typical for most benchmarks.
    
    \item \textit{Large} (100K--1M samples): Provides robust statistics and diverse coverage; requires automation.
    
    \item \textit{Very Large} (> 1M samples): Comprehensive coverage but may include redundancy; primarily for core capability assessment.
        
    \item \textit{Infinite}: Continuously generated evaluation data, enabling detection of overfitting to fixed test sets.
\end{itemize}

\paragraph{Data Organization.} Evaluations may also define \textit{splits} serving different purposes:

\begin{itemize}
    \item \textit{Fine-tuning/Development sets}: Used for task-specific model adaptation, requiring separation from evaluation data.
    
    \item \textit{Validation sets}: Public data for model selection and hyperparameter tuning, enabling fair comparison.
    
    \item \textit{Test sets}: Held-out data for final evaluation, ideally used sparingly to prevent indirect overfitting.
    
    \item \textit{Private/Hidden test sets}: Never-released data preventing direct optimization, though limiting reproducibility.
\end{itemize}

\paragraph{Temporal Characteristics.} The \textit{evaluation data type} captures temporal dynamics:

\begin{itemize}
    \item \textit{Static data}: Fixed test sets enabling direct comparison over time but vulnerable to contamination and saturation as models optimize for known benchmarks.
    
    \item \textit{Dynamic data}: Adaptive generation based on model responses, procedural creation, or periodic refresh. Prevents overfitting but complicates comparison across time.
    
    \item \textit{Composite}: Combining static and dynamic components.
\end{itemize}

\subsection{Method: Operational Procedures}
\label{subsec:method}

The \textbf{Method} dimension describes the technical implementation of evaluation: who or what judges outputs, what access to models is required, and how the evaluation protocol operates.

\paragraph{Judge Types.} The choice of evaluation judge represents one of the most consequential methodological decisions, directly affecting validity, cost, and scalability:

\begin{itemize}
    \item \textit{Human evaluation---Expert judges}: Domain specialists provide authoritative assessment of complex outputs (medical diagnoses, legal reasoning, creative writing). High validity but expensive, slow, and limited scale. Inter-rater reliability requires multiple experts per item.
    
    \item \textit{Human evaluation---Representative samples}: Crowdworkers or general population provide judgments aligned with typical users. More scalable than experts but requires careful quality control and may miss subtle errors.
    
    \item \textit{Automatic evaluation---Reference-based}: Comparing model outputs to ground truth using metrics like exact match, F1, BLEU, or learned similarity functions. Highly scalable and reproducible but validity depends on metric-target alignment.
    
    \item \textit{Automatic evaluation---Reference-free}: Assessing intrinsic properties without ground truth (perplexity, diversity, coherence scores). Enables evaluation when references are unavailable.
    
    \item \textit{Automatic evaluation---Execution-based}: Testing functional correctness by running code, executing actions in simulators, or checking logical consistency. Objective for well-defined tasks but limited to executable domains.
    
    \item \textit{Model-based evaluation---Expert models}: Fine-tuned or specialized models trained for evaluation (reward models, trained preference predictors). Can capture nuanced criteria but inherits biases from training data.
    
    \item \textit{Model-based evaluation---General LLMs}: Using powerful language models as judges. Highly scalable and applicable to open-ended tasks but introduces judge model biases, variance, and potential circularity when evaluating similar models.
    
    \item \textit{Hybrid evaluation}: Combining multiple judge types (e.g., automatic filtering followed by human evaluation of edge cases) to leverage complementary strengths.
\end{itemize}

For model-based judges, additional documentation should specify the judge model, prompting strategy, temperature settings, and measured inter-judge agreement when using multiple judge models or comparing to human judgments.

\paragraph{Evaluation Protocol.} The \textit{protocol} defines the step-by-step procedure for evaluation:

\begin{enumerate}
    \item Input selection and preprocessing
    \item Model prompting or query formatting
    \item Output generation parameters (temperature, top-p, max length)
    \item Post-processing and normalization
    \item Scoring or judgment procedures
    \item Aggregation across examples
    \item Statistical analysis and reporting
\end{enumerate}

Detailed protocol specification enables reproducibility and reveals potential confounds. For instance, whether examples are evaluated independently or model state persists across examples affects performance on context-dependent tasks.

\paragraph{Model Access Requirements.} Different evaluations require different levels of system access:

\begin{itemize}
    \item \textit{Output-only access}: Requires only the ability to query a model and observe outputs, applicable to commercial APIs where model internals are unavailable. Enables broad participation but limits evaluation to behavioral assessment.
    
    \item \textit{Partial access}: Requires intermediate representations, gradients, attention weights, or hidden states. Enables probing internal mechanisms and understanding model reasoning but unavailable for most production systems.
    
    \item \textit{Full access}: Requires complete access to architecture, weights, training procedures, and potentially training data. Enables comprehensive analysis including gradient-based robustness testing, interpretability analysis, and thorough contamination detection, but limited to open models.
\end{itemize}

Access requirements directly affect evaluation applicability---output-only evaluations apply broadly but partial or full access evaluations provide deeper insights for systems that permit such access.

\paragraph{Held-out Private Test Sets.} Many evaluations maintain \textit{private test sets} never released publicly to prevent direct optimization. Key considerations include:

\begin{itemize}
    \item \textit{Size and composition}: What fraction of data is held out? Is it representative of public splits?
    
    \item \textit{Access restrictions}: Is the private set used for periodic leaderboard updates, one-time challenges, or continuous evaluation services?
    
    \item \textit{Update frequency}: How often are private tests refreshed to prevent indirect optimization through repeated submissions?
    
    \item \textit{Purpose}: Preventing overfitting, detecting contamination, or ensuring fair comparison?
\end{itemize}

\subsection{Alignment: Reliability, Validity, and Robustness}
\label{subsec:alignment}

The \textbf{Alignment} dimension assesses whether an evaluation reliably measures what it claims to measure and whether conclusions drawn from results are warranted. Even well-designed evaluations can produce misleading results if sensitive to arbitrary choices, susceptible to contamination, or failing to measure claimed constructs.

\paragraph{Measurement Validation.} \textit{Measurement validation} establishes that an evaluation actually assesses its intended construct. Validation approaches include:

\begin{itemize}
    \item \textit{Expert review}: Domain specialists verify that evaluation items appropriately test target capabilities.
    
    \item \textit{Pilot studies}: Testing evaluation with known-good and known-bad systems to verify discriminative power.
    
    \item \textit{Correlation with established measures}: Demonstrating that new evaluation scores correlate with accepted benchmarks when measuring similar constructs.
    
    \item \textit{Ablation studies}: Removing evaluation components to verify they contribute to measuring target properties.
    
    \item \textit{Construct validity analysis}: Systematic examination of whether evaluation design aligns with theoretical understanding of measured capabilities.
\end{itemize}

Recent work by \citet{bean2025measuringmattersconstructvalidity} provides a comprehensive construct validity checklist for model evaluations, covering threat identification, measurement alignment, and validity argument construction. Evaluations meeting all checklist conditions provide stronger evidence for their claimed measurements.

\paragraph{Baselines and Points of Comparison.} Evaluation scores gain meaning through comparison. \textit{Baselines} provide reference points:

\begin{itemize}
    \item \textit{Random performance}: Chance-level accuracy establishes a lower bound.
    
    \item \textit{Simple heuristics}: Rule-based or retrieval systems demonstrate whether learning is necessary.
    
    \item \textit{Prior state-of-the-art}: Previous best results contextualize current performance.
    
    \item \textit{Human performance}: Expert or crowd performance indicates task difficulty and headroom for improvement, though human-model comparison requires careful consideration of different capabilities.
    
    \item \textit{Specialized models}: Domain-specific systems provide comparison points for general models.
\end{itemize}

\paragraph{Robustness Measures.} Evaluation robustness concerns whether results remain stable under reasonable variations. We distinguish several robustness dimensions:

\textit{Input robustness} captures sensitivity to prompt formatting, instruction phrasing, example ordering, or paraphrasing. Robust evaluations should yield consistent results for semantically equivalent inputs. Testing approaches include:
\begin{itemize}
    \item Multiple prompt templates for the same underlying task.
    \item Randomized example ordering.
    \item Paraphrased instructions preserving semantic content.
    \item Varied formatting (whitespace, capitalization, delimiters).
\end{itemize}

\textit{Output robustness} captures variability from model stochasticity. For probabilistic models, multiple evaluation runs with different random seeds reveal output variance. Reporting confidence intervals or standard deviations from multiple runs provides more complete performance characterization than single-run point estimates.

\textit{Evaluation procedure robustness} tests sensitivity to arbitrary methodological choices:
\begin{itemize}
    \item Multiple runs per sample with different random seeds
    \item Temperature and sampling parameter variations
    \item Different output length limits
    \item Repeated evaluations over time
\end{itemize}

\textit{Judge robustness} becomes critical for human or model-based evaluation. Measures include:
\begin{itemize}
    \item Inter-rater reliability (agreement between multiple human judges)
    \item Inter-model agreement (consistency across different judge models)
    \item Judge-human alignment (correlation between model judges and human assessments)
    \item Sensitivity to judge prompt variations
\end{itemize}

\textit{Statistical robustness} ensures conclusions don't depend on specific samples:
\begin{itemize}
    \item Confidence intervals and standard errors
    \item Statistical significance testing with appropriate corrections
    \item Effect size reporting (beyond just p-values)
    \item Bootstrapping or cross-validation for uncertainty quantification
\end{itemize}

\textit{Confound controls} verify that evaluation measures intended properties rather than spurious correlations:
\begin{itemize}
    \item Ablation studies removing supposedly critical components
    \item Negative controls that should not improve performance
    \item Minimal pair testing (inputs differing in single targeted aspects)
    \item Adversarial examples designed to exploit potential shortcuts
\end{itemize}

\paragraph{Known Limitations and Sensitivities.} Even after robustness testing, evaluations have \textit{known limitations}---documented sensitivities and failure modes that affect interpretation:

\begin{itemize}
    \item \textit{Format sensitivity}: "Performance varies ±5\% depending on prompt formatting"
    \item \textit{Domain specificity}: "Results may not generalize beyond formal written text"
    \item \textit{Demographic bias}: "Evaluation data underrepresents certain populations"
    \item \textit{Temporal validity}: "Evaluation uses 2023 knowledge cutoff; may not reflect current information"
    \item \textit{Self-preference bias}: "GPT-4 judge may favor outputs stylistically similar to GPT-4"
    \item \textit{Contamination risk}: "Public dataset may appear in training corpora"
\end{itemize}

\paragraph{Related Work and Differentiation.} Evaluations exist in an ecosystem of related benchmarks. \textit{Similar evaluations} should be documented along with key differences:

\begin{itemize}
    \item What prior benchmarks measure similar capabilities?
    \item How does this evaluation differ in scope, structure, or method?
    \item What specific gaps does this evaluation address?
    \item When should users prefer this evaluation over alternatives?
\end{itemize}

Clear positioning helps researchers select appropriate evaluation tools and understand how new evaluations advance the field.

\section{The \cardName\ Framework}
\label{sec:framework}

Operationalizing the taxonomy into a practical questionnaire requires balancing competing objectives: comprehensive coverage of all dimensions versus usability constraints, flexibility across evaluation types versus standardization for comparability, and mandatory requirements versus optional adaptability. We address these trade-offs through a structured design process that maps taxonomy categories to concrete questions while maintaining hierarchical organization.

\subsection{Design Principles}

Our framework adheres to six design principles derived from taxonomy requirements and anticipated use cases:

\begin{enumerate}[leftmargin=*]
\item \textbf{Comprehensiveness}: Cover all five taxonomy dimensions to ensure complete evaluation characterization (addresses transparency goal).

\item \textbf{Flexibility}: Accommodate diverse evaluation types through mandatory/optional question hierarchy (addresses adoption barrier).

\item \textbf{Accessibility}: Use plain language and provide extensive guidance (reduces completion burden).

\item \textbf{Actionability}: Generate concrete documentation enabling informed decisions (justifies completion effort).

\item \textbf{Comparability}: Maintain consistent structure across evaluation types (enables meta-analysis).

\item \textbf{Reproducibility}: Require sufficient detail for evaluation replication (supports scientific validity).
\end{enumerate}

These principles sometimes conflict---comprehensiveness versus accessibility, flexibility versus comparability. Our resolution strategy prioritizes mandatory elements addressing reproducibility while making comprehensive coverage optional, allowing users to balance documentation depth with available resources.

\subsection{Questionnaire Structure}

\paragraph{Derivation from Taxonomy.} Table~\ref{tab:taxonomy-mapping} shows the mapping between taxonomy dimensions and questionnaire sections. Each taxonomy category generates at least one question, with complex categories expanding into multiple questions or sub-questions. For instance, Context (\S\ref{subsec:context}) maps to the Basic Information section with questions about provenance, authorship, release date, and purpose; Scope (\S\ref{subsec:scope}) maps to the What Does It Evaluate section with questions about capabilities tested, model properties evaluated, and input/output modalities; Structure (\S\ref{subsec:structure}) maps to the How Is It Structured section with questions about data sources, size, splits, and design; Method (\S\ref{subsec:method}) maps to the How Does It Work section with questions about judge type, evaluation protocol, and model access requirements; and Alignment (\S\ref{subsec:alignment}) maps to the Quality \& Reliability section with questions about measurement validation, baselines, robustness measures, and known limitations.

\begin{table}[h]
\centering
\caption{Mapping between taxonomy categories and questionnaire sections}
\label{tab:taxonomy-mapping}
\begin{tabular}{ll}
\toprule
\textbf{Taxonomy Category} & \textbf{Questionnaire Section(s)} \\
\midrule
Context & Basic Information \\
Scope & What Does It Evaluate \\
Structure & How Is It Structured \\
Method & How Does It Work \\
Alignment & Quality \& Reliability \\
\bottomrule
\end{tabular}
\end{table}

\subsection{Integration with Existing Frameworks}

\cardNamePlural\ complements existing documentation frameworks through composition rather than replacement. Datasheets document datasets used in evaluations (referenced in the How Is It Structured section), Model Cards document systems being evaluated (referenced in the Basic Information section), and System Cards document complete deployments including evaluation results (include \cardNamePlural\ as a component).

A complete transparency ecosystem involves all four frameworks. Model Card documents the model, Datasheet documents training data, and \cardNamePlural\ document each assessment methodology. This modular approach enables reuse---a single \cardName\ can be referenced by multiple Model Cards or System Cards, avoiding duplication while maintaining comprehensive documentation.

\section{Case Studies}
\label{sec:examples}

To validate that \cardNamePlural\ effectively captures diverse evaluation paradigms, we present three case studies spanning traditional benchmarks (ImageNet~\citep{imagenet}), execution-based testing (HumanEval~\citep{humaneval}), and emerging LLM-as-judge methodologies (MT-Bench~\citep{mtbench}). Each case study demonstrates how the unified questionnaire structure accommodates paradigm-specific characteristics while maintaining consistency and comparability.

\subsection{Example 1: ImageNet (Computer Vision Benchmark)}

\begin{evaluationcard}[
  title={ImageNet},
  subtitle={ImageNet enables large-scale visual recognition research and provides a standardized benchmark for comparing computer vision models across diverse, hierarchically-organized object categories.},
  authors={Priceton University},
  link={https://ieeexplore.ieee.org/document/5206848},
  date={2009}
]

  \Purpose{Research; Model Selection}
  \PrinciplesTested{Object Recognition; Visual Understanding}
  \FunctionalProps{General Capability (object recognition)}
  \InputModality{Images}
  \OutputModality{Class predictions}

  \InputSource{Curated dataset (publicly available)}
  \OutputSource{Human annotated}
  \Size{Large (>100K samples): 14 million images total}
  \Splits{Development set: 1.2M images

Validation set: 50K images

Test set: 100K images (labels withheld on server)}
  \Design{Static benchmark}

  \Judge{Automatic (Reference-based)}
  \Protocol{1) Model receives image
2) Model outputs class predictions
3) Automatic scoring against ground truth labels}
  \ModelAccess{Black-box (outputs only)}
  \HasHeldout{true}
  \HeldoutDetails{Test set: 100K images with labels withheld on evaluation server}

  \AlignmentValidation{Human performance establishes ceiling: 94.9\% top-5 accuracy. Images organized hierarchically using WordNet taxonomy.}
  \RobustnessMeasures{Extensive annotation procedures; Multiple validation rounds}
  \KnownLimitations{Label noise (~5\% estimated error rate); Geographic and cultural bias toward Western contexts; Outdated categories; Test set saturation and contamination concerns; Not suitable for fine-grained recognition, medical imaging (domain shift), or fairness assessment}
  \BenchmarksList{ImageNet Large Scale Visual Recognition Challenge (ILSVRC)}

\end{evaluationcard}

\subsection{Example 2: HumanEval (Code Generation Benchmark)}

\begin{evaluationcard}[
  title={HumanEval},
  subtitle={HumanEval evaluates functional correctness of code generated by language models through execution-based testing, addressing gaps in existing evaluations that focused on code similarity rather than correctness.},
  authors={OpenAI},
  link={https://arxiv.org/abs/2107.03374},
  code-link={https://github.com/openai/human-eval},
  date={2021}
]

  \Purpose{Research}
  \PrinciplesTested{Algorithmic problem-solving; Language-specific syntax; Edge case handling}
  \FunctionalProps{General Capability (code generation); Correctness (passes unit tests)}
  \InputModality{Text (programming problems)}
  \OutputModality{Code (Python functions)}

  \InputSource{New dataset (released with eval)}
  \OutputSource{Programmatically generated (unit tests)}
  \Design{Static benchmark}

  \Judge{Automatic (Execution-based)}
  \Protocol{1) Model generates function implementation from problem description

2) Generated code is executed against unit tests

3) Scored as pass/fail per test}
  \ModelAccess{Outputs}
  \HasHeldout{false}
  \HeldoutDetails{N/A - test cases are public}

  \AlignmentValidation{Problems hand-written by experienced programmers. Each problem verified with multiple test cases covering edge cases.}
  \BaselineModels{Performance varies by model; prompt formatting affects performance by ~10\%}
  \RobustnessMeasures{Minimal pairs tested for similar problem variations; Multiple test cases per problem}
  \KnownLimitations{Python only; Only algorithmic problems (no system design); Small-scale (164 problems); Sensitive to prompt formatting (instructional vs. completion style); Test cases are public, enabling contamination; Not suitable for production code quality assessment (no tests for style, documentation, security), multi-language comparison, or complex system design}
  \BenchmarksList{HumanEval+ (extended test suite); MultiPL-E (multi-language variant)}

\end{evaluationcard}

\subsection{Example 3: MT-Bench (Conversational AI Arena)}

\begin{evaluationcard}[
  title={MT-Bench},
  subtitle={MT-Bench evaluates multi-turn conversational ability using LLM-as-judge methodology, addressing limitations of single-turn benchmarks that don't capture dialogue coherence and context-tracking.},
  authors={UC Berkeley},
  link={https://arxiv.org/abs/2306.05685},
  code-link={https://github.com/lm-sys/FastChat/tree/main/fastchat/llm_judge\#mt-bench/},
  date={2023}
]

  \Purpose{Research}
  \PrinciplesTested{Conversational coherence; Instruction following; Multi-turn context tracking}
  \FunctionalProps{General Capability (conversation); Quality (helpfulness, coherence)}
  \InputModality{Text (multi-turn questions)}
  \OutputModality{Text (conversational responses)}

  \InputSource{Curated dataset}
  \OutputSource{Human annotated}
  \Size{Small (<1K samples): 80 multi-turn questions across 8 categories}
  \Splits{Each question has 2 turns
Categories: writing, roleplay, reasoning, math, coding, extraction, STEM, humanities}
  \Design{Dynamic data-driven (continuous arena with periodic updates)}

  \Judge{Model-based (LLM judge: GPT-4)}
  \Protocol{1) Model A and Model B answer same question
2) GPT-4 compares responses pairwise
3) Winner determined; Elo rating computed from pairwise comparisons}
  \ModelAccess{Outputs}
  \HasHeldout{false}
  \HeldoutDetails{Questions periodically updated to prevent memorization}

  \AlignmentValidation{GPT-4 judgments correlate with human preferences at 80\% agreement rate. Questions carefully crafted to avoid common knowledge.}
  \BaselineModels{Elo ratings computed from pairwise comparisons; Public leaderboard updated weekly}
  \RobustnessMeasures{Position bias controlled via swapping; Inter-judge reliability: Claude-2 shows 75\% agreement with GPT-4; Prompt sensitivity tested: stable across 3 judge prompt variations}
  \KnownLimitations{Judge model bias (prefers certain styles); Limited to English; Small question set may enable memorization; Not suitable for domain-specific expertise evaluation, safety assessment (doesn't test for harmful outputs), or fine-grained capability measurement (coarse Elo ratings)}
  \BenchmarksList{Chatbot Arena, AlpacaEval}

\end{evaluationcard}

\section{Conclusion}

We introduced \cardNamePlural\, a structured framework for documenting AI evaluation methodologies through a five-dimensional taxonomy and comprehensive questionnaire. Our framework addresses a critical gap in AI transparency: while datasets and models benefit from standardized documentation, evaluation methodologies---equally important for understanding AI capabilities and limitations---have lacked systematic documentation standards.

Through three diverse case studies (ImageNet, HumanEval, MT-Bench), we demonstrated that \cardNamePlural\ effectively captures evaluations spanning traditional benchmarks, execution-based testing, and emerging LLM-as-judge paradigms. The \cardName\ environment provides a standardized format that accommodates paradigm-specific requirements while maintaining consistent structure for comparability.

Key contributions include: (1) a systematic taxonomy of evaluation characteristics organized across five orthogonal dimensions---Context, Scope, Structure, Method, and Alignment, (2) a practical questionnaire operationalizing this taxonomy through structured sections that map directly to these dimensions, balancing comprehensiveness and usability, and (3) validated examples in the \texttt{evaluationcard} format demonstrating applicability across diverse evaluation types.

We encourage the community to create \cardNamePlural\ for their methodologies, contribute domain-specific extensions, and provide feedback to inform framework evolution. By establishing documentation standards for evaluation, we can improve transparency, reproducibility, and informed decision-making.

\bibliographystyle{assets/plainnat}  
\bibliography{bib}



\end{document}